\documentclass[hidelinks]{article}
\pdfoutput=1

\usepackage[preprint]{neurips_2023}

\usepackage{booktabs, graphicx, subcaption, pgfplots}
\usepackage[inline]{enumitem}
\usepackage[utf8]{inputenc} 
\usepackage[T1]{fontenc}    
\usepackage{hyperref}       
\usepackage{url}            
\usepackage{booktabs}       
\usepackage{amsfonts}       
\usepackage{nicefrac}       
\usepackage{microtype}      
\usepackage{xcolor}         

\usepackage{natbib}

\title{Does Synthetic Data Make Large Language Models More Efficient?}

\author{
    Sia Gholami \\
    The Institute of Electrical and Electronics Engineers, Member IEEE \\
    \texttt{gholami@ieee.org} \\
    \And 
    Marwan Omar \\
    Illinois Institute of Technology \\
    \texttt{momar3@iit.edu} \\
}

\begin{document}

\maketitle

\begin{abstract}
Natural Language Processing (NLP) has undergone transformative changes with the advent of deep learning methodologies. One challenge persistently confronting researchers is the scarcity of high-quality, annotated datasets that drive these models. This paper explores the nuances of synthetic data generation in NLP, with a focal point on template-based question generation. By assessing its advantages, including data augmentation potential and the introduction of structured variety, we juxtapose these benefits against inherent limitations, such as the risk of overfitting and the constraints posed by pre-defined templates. Drawing from empirical evaluations, we demonstrate the impact of template-based synthetic data on the performance of modern transformer models. We conclude by emphasizing the delicate balance required between synthetic and real-world data, and the future trajectories of integrating synthetic data in model training pipelines. The findings aim to guide NLP practitioners in harnessing synthetic data's potential, ensuring optimal model performance in diverse applications.
\end{abstract}

\section{Introduction}
In the burgeoning field of Natural Language Processing (NLP), acquiring substantial data for training and fine-tuning models is a continual challenge~\citep{vaswani2017attention}. While real-world annotated datasets are invaluable, their availability is often constrained, making them expensive to produce, and they can sometimes carry inherent biases from their collection methods~\citep{bowman2015large}. This context underscores the potential of synthetic data generation techniques, with synthetic question-answer pairs emerging as a notable subset~\citep{rajpurkar2016squad}. Among the diverse strategies available, template-based question generation, recognized for its rule-driven approach, provides a systematic avenue for data generation~\citep{chen2016thorough}.

However, as with many techniques in the realm of computational linguistics, the adoption of template-based generation in transformer models within NLP presents a complex landscape to navigate~\citep{devlin2018bert}. This paper seeks to illuminate the intricacies of this approach, offering insights into its methodologies, advantages, and inherent limitations. Through our examination, our aim is to equip readers with a nuanced understanding of the technique, its impact on transformer architectures, and the potential avenues for its evolution in NLP research.


The implementation of template-based question generation for creating synthetic question-answer pairs can significantly impact the performance of a LLM in several ways:

\begin{enumerate}
    \item Data Augmentation: The most direct impact is the increase in training data. When you create synthetic question-answer pairs from existing text, you're effectively augmenting your dataset, which can be particularly useful when dealing with tasks where the amount of available labeled data is limited. This increased data volume helps the model better understand language patterns and variations, which can enhance the model's ability to generalize, ultimately improving performance.

    \item Exposure to Diverse Structures: Template-based question generation exposes the transformer model to a wider variety of question structures and types. This increased exposure helps the model develop a more comprehensive understanding of language and better performance on a broader range of questions.

    \item Model Robustness: By creating synthetic data that includes a variety of linguistic features and structures, the model becomes more robust. It will be less likely to overfit to the training data, and it will perform better when encountering previously unseen data, increasing its robustness and reliability.

    \item Bias Mitigation: Synthetic data can help to mitigate biases in the original dataset by introducing more balanced and diverse examples. This can make the model's predictions less skewed and more reliable.
\end{enumerate}

However, it's important to note that while these potential benefits are significant, they are not guaranteed. The quality of the synthetic question-answer pairs is crucial. If the generated synthetic data is of low quality or doesn't accurately reflect the kinds of questions and answers the model will encounter in the real world, it might instead negatively impact the model's performance~\citep{kim2019improving}.

Moreover, while template-based question generation can create a diverse range of questions, it's inherently limited by the predefined templates. Therefore, it may not capture all possible ways of phrasing questions or handling complex sentence structures. For these reasons, template-based generation is often used in conjunction with other question generation methods or with fine-tuning on real-world data to ensure that the transformer model is well-prepared for the task at hand.

\section{Related Works}
Natural Language Processing (NLP) has been a major area of research in Artificial Intelligence and Machine Learning
since the early days of computer science~\citep{voorhees1999trec, moldovan2000structure, brill2002analysis, ferrucci2010building, gholami2021zero, gholami2022you, gholami2022create, gholami2022alexa, gholami2022flight, brand2022text, gholami2023generative, gholami2023student, gholami2023pruning}. There are numerous works of leveraging synthetic data to create efficient Transformer models in the literature. In this section we go over a few notable cases.

In the real world, there is plenty of unlabeled data. However, it could be challenging to locate task-specific unlabeled data that fits the criteria of a particular machine-learning scenario. In particular, it is challenging to locate in-domain unlabeled text that complies with a particular Natural Language Processing system's probability model. To produce an enhanced subset of features, additional data is often included with the existing training sample in classical data augmentation. Furthermore, labeled ambiguity can harm training if combined with data generated using a generative model. Additionally, the created queries may need to be more logical and clear noise. Yang et al. ~\citep{yang2020generative} addressed the problem by providing a straightforward training method that handles natural and synthetic information separately. They initially build a model by using artificial data and afterward refine it using the original, human-created training dataset. For computer vision, dataset augmentation frequently leads to the formation of visual modifications like translational and rotational. 

Data augmentation is more difficult for language applications. Back-translation configurations, heuristic analysis based on text's semantic and syntactic characteristics, such as phrase replacement options using a word list, and more lately, generative algorithms for replicating new and more effective instances for character recognition and reading ability, have all generally been employed in preceding experimental tools. To enhance the functionality of detectors, Tavor et al. ~\citep{anaby2020not}  present the LAMBADA learning algorithm is adjusted and created additional labeled-condition phrases involving the filtering step. They demonstrated that their approach significantly enhances classifiers' performance on smaller data sets. Furthermore, they demonstrated that LAMBADA outperforms cutting-edge methods for data augmentation.
Alberti et al. ~\citep{alberti2019synthetic} introduced a new technique for creating synthetic Question Answer examples and showed how this information improved SQuAD2 as well as NQ. Furthermore, they suggested a potential course of action for this methodology's logical foundation, which will be explored further in later studies.

Several new techniques for synthetic data production analysis of large pretrained language algorithms have begun to show results in enhancing the progress of the Reading Cognition test with artificially generated data. Given the limited amount of human-labeled data, a set of questions and their answers creation is a data augmentation technique used to enhance question-answering (QA) frameworks. In order to develop a BERT-Large model to attain comparable question-answering efficiency while explicitly utilizing any actual information, Puri et al. ~\citep{puri2020training} constructed artificial content using a Wikipedia-fine tuned GPT-2 system that generates response alternatives as well as artificial queries dependent upon these responses.

\section{Approach}
In this section we propose a method for generating synthetic question-answer pairs. Creating synthetic question-answer pairs from a text corpus requires an in-depth understanding of the text content and a detailed mapping of its semantic and syntactic structure. Here's a more detailed description of the process:

\begin{enumerate}
    \item Preprocessing: Preprocessing involves cleaning and standardizing the text corpus. This includes tasks like removing punctuation, lowercasing text, expanding contractions, correcting spelling, and so on. This step prepares the text for further processing and analysis.

    \item Sentence Segmentation: Sentence segmentation, or sentence boundary detection, is the process of splitting a text into individual sentences. Each sentence can then be analyzed separately for the generation of question-answer pairs.

    \item Parsing and Text Analysis: 
    \begin{itemize}
        \item Part-of-Speech Tagging: This process assigns each word in the sentence its respective part of speech (such as noun, verb, adjective, etc.), based on its context and definition.
        \item Named Entity Recognition (NER): NER locates and classifies named entities in text into predefined categories like persons, organizations, locations, etc.
        \item Dependency Parsing: Dependency parsing analyzes the grammatical structure of a sentence, establishing relationships between words, and determining how words relate to each other.
   \end{itemize}

    \item Template-based Question Generation: Using predefined templates for different question types (who, what, when, where, why, how), questions are generated based on the entities and relationships found in the text. For instance, if a sentence mentions a specific event happening at a specific time, a "when" question can be formulated.
    
    \item Answer Extraction: For every generated question, the corresponding answer is the segment or specific detail from the original text that the question was based on. This can range from a single word or phrase to a whole sentence or more.

    \item Training a Model: The generated synthetic question-answer pairs can then be used to train a Question Answering (QA) model. This is often a supervised learning task, where the model learns to predict the answer given a question and context. Transformer models like BERT or T5 are commonly used for this task due to their effectiveness in understanding context and extracting relevant information.

    \item Evaluation and Refinement: Finally, the model's performance is evaluated, ideally on a separate test set of question-answer pairs. The synthetic data generation process and the model can be iteratively refined based on the model's performance and any observed shortcomings.
\end{enumerate}

Generating high-quality synthetic question-answer pairs is a complex task that requires careful design and refinement of the question generation and answer extraction processes. However, when done effectively, it can significantly enhance the performance of LLMs, especially when real-world annotated QA datasets are scarce or unavailable.

In this technique, predefined templates for different question types (like who, what, when, where, why, how) are used, which are then filled with appropriate information extracted from the source text to generate relevant questions. Here are the steps: 

\begin{enumerate}
    \item Identify Suitable Sentences: The first step in template-based question generation involves identifying sentences in the text that contain the potential to form meaningful questions. This might involve looking for sentences with clear subjects, objects, and verbs, or sentences containing named entities (people, places, dates, etc.) or interesting facts.

    \item  Extract Key Information: The next step involves extracting key pieces of information from the identified sentences. This typically involves applying techniques like Named Entity Recognition (NER) to identify key entities, dependency parsing to understand the sentence structure, and part-of-speech tagging to understand the role of each word in the sentence.

    \item Apply Templates: Once the key information is extracted from a sentence, it is inserted into a suitable question template. Templates are predefined structures of questions, designed to cover common question forms. For instance, templates might include structures like:
    \begin{itemize}
        \item[--] "Who [verb] [object]?"
        \item[--] "What [verb] [subject]?"
        \item[--] "When did [subject] [verb]?"
        \item[--] "Where is [object]?"
    \end{itemize}
    The specific template chosen depends on the type and structure of the information extracted from the sentence. For example, if the sentence mentions a person doing an action, the "Who [verb] [object]?" template might be used.

    \item Refine Questions: After initial question generation, the questions might be refined to improve readability, correct grammar errors, or ensure they make sense in the context of the text. This might involve minor text edits or rerunning the question generation process with different templates.
\end{enumerate}

In this study, we focus on adding synthetic data to the model introduced by~\cite{gholami2023generative} (GPT-Efficio) as the baseline along with bigger GPT-3~\citep{brown2020language} model.

While template-based question generation can be a powerful tool for creating synthetic question-answer pairs, it does have limitations. It's typically rule-based, meaning it may struggle with complex or ambiguous sentences that don't fit neatly into its predefined templates. Moreover, the diversity of the generated questions is limited to the predefined templates. This is why more advanced, machine-learning-based question generation techniques are also used, often in conjunction with template-based methods, to generate a wider range of question types and handle more complex sentence structures. 

To overcome these limitations, modern approaches often employ transformer-based models or sequence-to-sequence models that are cble of learning the complex mappings from source sentences to questions from large amounts of training data. Nevertheless, template-based question generation still plays a crucial role, particularly in scenarios with limited data or where interpretability and control over the generation process are important.

\section{Experiments}
By artificially creating data that closely mimics genuine datasets, the potential to enrich training sets and address data scarcity becomes tangible. Yet, as with all innovations, its efficacy is contingent on context and application.

For language modeling tasks, synthetic data generation might appear as a beacon of promise on the surface. Here we have a chance to artificially bolster the data pool, potentially leading to better-trained models capable of understanding and predicting linguistic structures. However, the reality reveals a different narrative. The inherent nature of language modeling, where the task revolves around predicting subsequent words in sentences or deciphering intricate linguistic patterns, demands a nuanced and authentic representation of the language. Synthetic data, even when finely crafted, may not capture the intricate unpredictability and vastness of natural language. Consequently, its inclusion often results in minimal to negligible improvements in model accuracy and fluency. This could be attributed to various factors, including the potential for synthetic data to introduce noise or fail to capture the linguistic variances found in genuine, human-generated text.

On the contrary, when examining question generation tasks, synthetic data generation has shown to be of greater relevance. Unlike the broad scope of language modeling, question generation is more constrained, relying on structured formats and specific linguistic cues. Given its rule-based nature, synthetic data can be tailored to this task more effectively, providing models with a plethora of varied question formats and structures. Our investigations indicate that, while the improvements might not be groundbreaking, there is a discernible enhancement in the model's ability to generate coherent and relevant questions when trained with a blend of real and synthetic data. It's possible that the structured nature of questions allows synthetic generation techniques to produce data that is more aligned with the inherent patterns of question formulation, hence the observed performance boost.

\subsection{Results}

\begin{table}[!htbp]
\centering
\small
\caption{Performance of synthetic question-answer generation on completion tasks}\label{syn-lm}
\begin{tabular}{p{0.22\linewidth} p{0.07\linewidth} p{0.15\linewidth} p{0.15\linewidth} p{0.12\linewidth} p{0.12\linewidth}}
\toprule
\textbf{Model} & \textbf{$n_{params}$} & \textbf{LAMBADA (acc)} & \textbf{LAMBADA (ppl)} & \textbf{StoryCloze (acc)} & \textbf{HellaSwag (acc)} \\ 
\midrule
GPT-3 Zero-Shot & 175B & 76.2 & 3.00 & 83.2 & 78.9   \\ 
GPT-3 One-Shot & 175B & 72.5 & 3.35 & 84.7 & 78.1   \\ 
GPT-3 Few-Shot & 175B & 86.4 & 1.92 & 87.7 & 79.3   \\ 
GPT-Efficio & 950M & 67.1 & 9.2 & 80.5 & 72.6 \\
GPT-Efficio (+ synQA) & 950M & 67.1 & 9.2 & 80.5 & 72.6 \\
\bottomrule
\end{tabular}
\end{table}

Table~\ref{syn-lm} demonstrates the GPT-Efficio performance with and without synthetic data in comparison with GPT-3 in language modeling tasks.

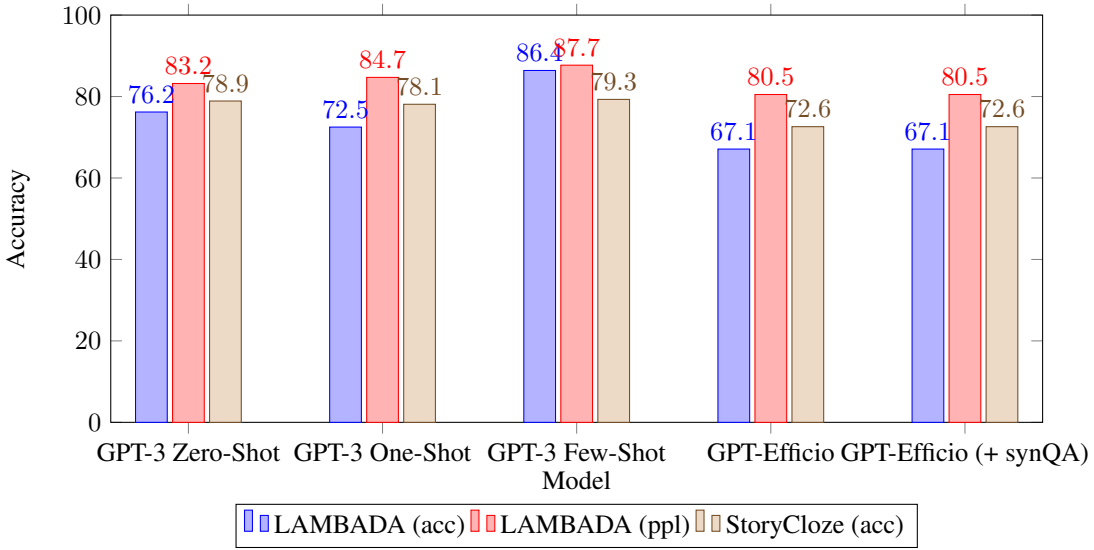
\begin{figure}[htbp]
\centering
\begin{tikzpicture}
\begin{axis}[
    ybar, ymin=0, ymax=100,
    width=\textwidth,
    height=7cm,
    xlabel={Model},
    ylabel={Accuracy},
    symbolic x coords={GPT-3 Zero-Shot, GPT-3 One-Shot, GPT-3 Few-Shot, GPT-Efficio, GPT-Efficio (+ synQA)},
    xtick=data,
    legend style={at={(0.5,-0.2)}, anchor=north,legend columns=-1},
    ymin=0,
    ymax=100,
    nodes near coords,
    nodes near coords align={vertical},
    bar width=12pt,
    ]
    
\addplot coordinates {(GPT-3 Zero-Shot, 76.2) (GPT-3 One-Shot, 72.5) (GPT-3 Few-Shot, 86.4) (GPT-Efficio, 67.1) (GPT-Efficio (+ synQA), 67.1)};
\addplot coordinates {(GPT-3 Zero-Shot, 83.2) (GPT-3 One-Shot, 84.7) (GPT-3 Few-Shot, 87.7) (GPT-Efficio, 80.5) (GPT-Efficio (+ synQA), 80.5)};
\addplot coordinates {(GPT-3 Zero-Shot, 78.9) (GPT-3 One-Shot, 78.1) (GPT-3 Few-Shot, 79.3) (GPT-Efficio, 72.6) (GPT-Efficio (+ synQA), 72.6)};

\legend{LAMBADA (acc), LAMBADA (ppl), StoryCloze (acc), HellaSwag (acc)}

\end{axis}
\end{tikzpicture}
\caption{Performance of synthetic question-answer generation on completion tasks}\label{fig:syn-lm}
\end{figure}

\begin{table}[!htbp]
\centering
\small
\caption{Performance of synthetic question-answer on QA tasks}\label{syn-qa}
\begin{tabular}{p{0.23\linewidth} p{0.12\linewidth} p{0.12\linewidth} p{0.12\linewidth} p{0.12\linewidth}}
\toprule
\textbf{Model} & \textbf{$n_{params}$} & \textbf{NQ} & \textbf{WebQ} & \textbf{TriviaQA}\\ 
\midrule
GPT-3 Zero-Shot & 175B & 14.6 & 14.4 & 64.3   \\ 
GPT-3 One-Shot & 175B & 23.0 & 25.3 & 68.0   \\ 
GPT-3 Few-Shot & 175B & 29.9 & 41.5 & 71.2   \\ 
GPT-Efficio & 950M & 27.5 & 40.6 & 69.2 \\
GPT-Efficio (+ synQA) & 950M & 28.43 & 42.12 & 70.45 \\
\bottomrule
\end{tabular}
\end{table}

Table~\ref{syn-qa} shows the GPT-Efficio performance with and without synthetic data in comparison with GPT-3 in question answering tasks.

\begin{figure}[!htbp]
\centering
\begin{tikzpicture}
\begin{axis}[
    ybar, ymin=0, ymax=100,
    bar width=0.2cm,
    width=12cm,
    height=8cm,
    xlabel={Model},
    ylabel={Accuracy},
    symbolic x coords={GPT-3 Zero-Shot,GPT-3 One-Shot,GPT-3 Few-Shot,GPT-Efficio,GPT-Efficio (+ synQA)},
    xtick=data,
    legend style={at={(0.5,-0.15)},
    x tick label style={rotate=25,anchor=east},
      anchor=north,legend columns=-1},
    ymin=0,
    ymax=80,
    nodes near coords,
    nodes near coords align={vertical},
    every node near coord/.append style={rotate=90, anchor=west, font=\footnotesize},
    legend style={at={(1,1)},
        anchor=south east,legend columns=-1}
    ]
\addplot coordinates {(GPT-3 Zero-Shot,14.6) (GPT-3 One-Shot,23.0) (GPT-3 Few-Shot,29.9) (GPT-Efficio,27.5) (GPT-Efficio (+ synQA),28.43)};
\addplot coordinates {(GPT-3 Zero-Shot,14.4) (GPT-3 One-Shot,25.3) (GPT-3 Few-Shot,41.5) (GPT-Efficio,40.6) (GPT-Efficio (+ synQA),42.12)};
\addplot coordinates {(GPT-3 Zero-Shot,64.3) (GPT-3 One-Shot,68.0) (GPT-3 Few-Shot,71.2) (GPT-Efficio,69.2) (GPT-Efficio (+ synQA),70.45)};
\legend{NQ, WebQ, TriviaQA}
\end{axis}
\end{tikzpicture}
\caption{Performance of synthetic question-answer on QA tasks}
\label{fig:syn-qa}
\end{figure}
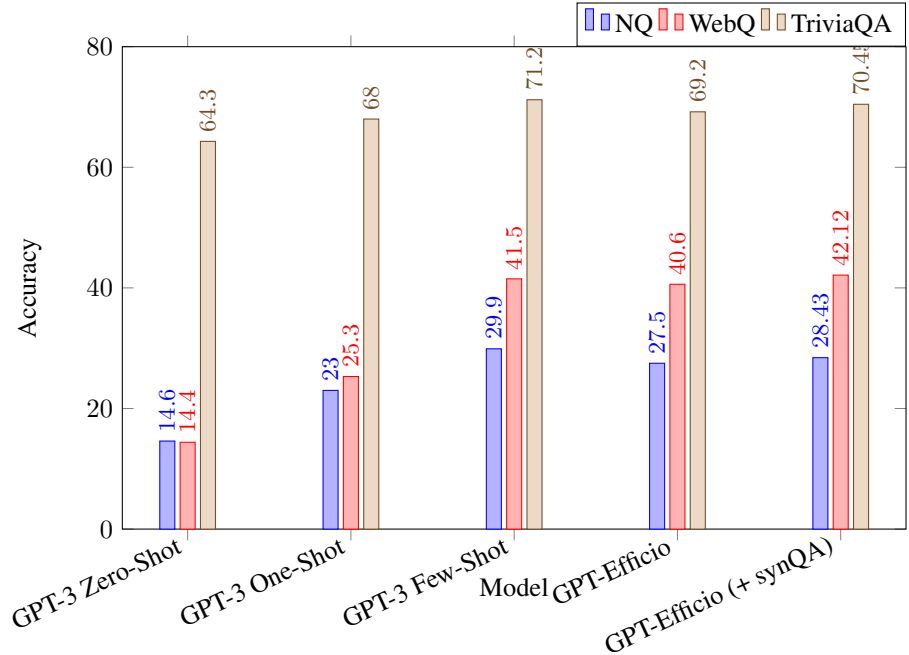

\section{Analysis}
Hyperparameters in the context of template-based synthetic question generation are related to the construction and selection of templates, and how data is processed for template filling. Hyperparameters include:

\begin{enumerate} 
    \item Number of Templates: This refers to the total number of different question templates used. Too few templates could limit the diversity of questions, making the model less robust to different question formulations. Too many, and the model might spread its learning too thin, struggling to learn any particular pattern well.

    \item Template Complexity: This refers to the complexity of the templates in terms of their linguistic structures. Simpler templates could make the learning process easier, but might limit the ability of the model to handle more complex sentences.~\citep{mariotti2020towards}. More complex templates can help the model handle a wider range of sentence structures, but may also make the learning process more challenging.

    \item Entity and Relationship Extraction Parameters: These could include parameters related to how entities and relationships are extracted from sentences for filling in the templates. This could involve the thresholds used to decide when a particular word or phrase is considered an entity or part of a relationship.

    \item Threshold for Question Selection: Not every generated question will be of high quality. Some threshold or criteria might be set to determine which questions are included in the final synthetic dataset.~\citep{bao2018question}.

    \item Ratio of Synthetic to Real Data: If synthetic data is being combined with real data, the ratio of synthetic to real data used could significantly impact the model's performance. Too much synthetic data could lead the model to overfit to the patterns in the synthetic data and perform poorly on real data.
\end{enumerate}
The effects of these hyperparameters on the performance of a Language Learning Model (LLM) can vary widely depending on the specific implementation and application. Generally, they would affect the quality and diversity of the synthetic question-answer pairs generated, and therefore the amount and type of information the model can learn from. Adjusting these hyperparameters should be done carefully, with consideration for the specific learning task and based on validation performance, to ensure the best possible performance of the LLM.

In this section we focus on the ratio of synthetic to real data hyperparameters. The ratio of synthetic to real data is a significant hyperparameter in the training of language models when using synthetic data. It refers to the proportion of synthetic data samples versus real (or naturally occurring) data samples in your training dataset.~\citep{sennrich2015improving}.

When creating the training dataset, a few factors come into play:

\begin{enumerate}
    \item Quality of Synthetic Data: The quality of your synthetic data plays a crucial role in determining an optimal ratio. If the synthetic data is of high quality, closely mirroring the statistical properties of real-world data, then a higher ratio of synthetic to real data might be beneficial. On the other hand, if the synthetic data is of lower quality or does not represent the real-world distribution well, a lower ratio is usually better to avoid the model overfitting to the synthetic data's characteristics.

    \item Size of Original Dataset: If the original dataset is small, adding a substantial amount of synthetic data can help to augment the dataset, leading to better model performance due to increased diversity and quantity of training samples.

    \item Task Complexity: For complex tasks that require understanding of nuanced language use, too high a ratio of synthetic to real data could harm performance, since synthetic data might not fully capture these nuances. 
\end{enumerate}
The ratio of synthetic to real data affects the training in various ways:

\begin{itemize}
    \item Positive Effects: Increasing the proportion of synthetic data can help in data augmentation, effectively increasing the size of your training dataset. This can be particularly useful when dealing with tasks where the amount of available labeled data is limited. It can help expose the model to a wider variety of scenarios and edge cases, making the model more robust.~\citep{brown2020language}.

    \item Negative Effects: If the synthetic data doesn't well represent the distribution of real data, having too much synthetic data can cause the model to learn patterns that don't generalize well to real data. This is a form of overfitting, where the model performs well on the training data but poorly on unseen, real-world data.
\end{itemize}
Determining the right balance typically involves empirical testing. Starting with a lower ratio of synthetic to real data and gradually increasing it, monitoring the model's performance on a validation dataset. A good strategy is to use cross-validation or a hold-out validation set to tune this hyperparameter, similar to other forms of hyperparameter tuning in machine learning~\citep{dathathri2019plug}. This approach can help ensure that the chosen ratio leads to the best possible model performance.

\begin{table}[!htbp]
\centering
\small
\caption{Analysis of the effects of hyperparameter synthetic to real data rate on completion tasks}\label{syn-anal-lm}
\begin{tabular}{p{0.18\linewidth} p{0.06\linewidth} p{0.06\linewidth} p{0.15\linewidth} p{0.15\linewidth} p{0.12\linewidth} p{0.12\linewidth}}
\toprule
\textbf{Model} & \textbf{syn\%} & \textbf{$n_{params}$} & \textbf{LAMBADA (acc)} & \textbf{LAMBADA (ppl)} & \textbf{StoryCloze (acc)} & \textbf{HellaSwag (acc)} \\ 
\midrule
GPT-3 Zero-Shot & - & 175B & 76.2 & 3.00 & 83.2 & 78.9   \\ 
GPT-3 One-Shot & - & 175B & 72.5 & 3.35 & 84.7 & 78.1   \\ 
GPT-3 Few-Shot & - & 175B & 86.4 & 1.92 & 87.7 & 79.3   \\ 
GPT-Efficio & - & 950M & 67.1 & 9.2 & 80.5 & 72.6 \\
GPT-Efficio & .1 & 950M & 67.1 & 9.2 & 80.5 & 72.6 \\
GPT-Efficio & .3 & 950M & 67.1 & 9.2 & 80.5 & 72.6 \\
GPT-Efficio & .5 & 950M & 67.11 & 9.2 & 80.53 & 72.62 \\
\bottomrule
\end{tabular}
\end{table}

Table~\ref{syn-anal-lm} demonstrates the GPT-Efficio performance with and without synthetic data in comparison with GPT-3 in language modeling tasks. 
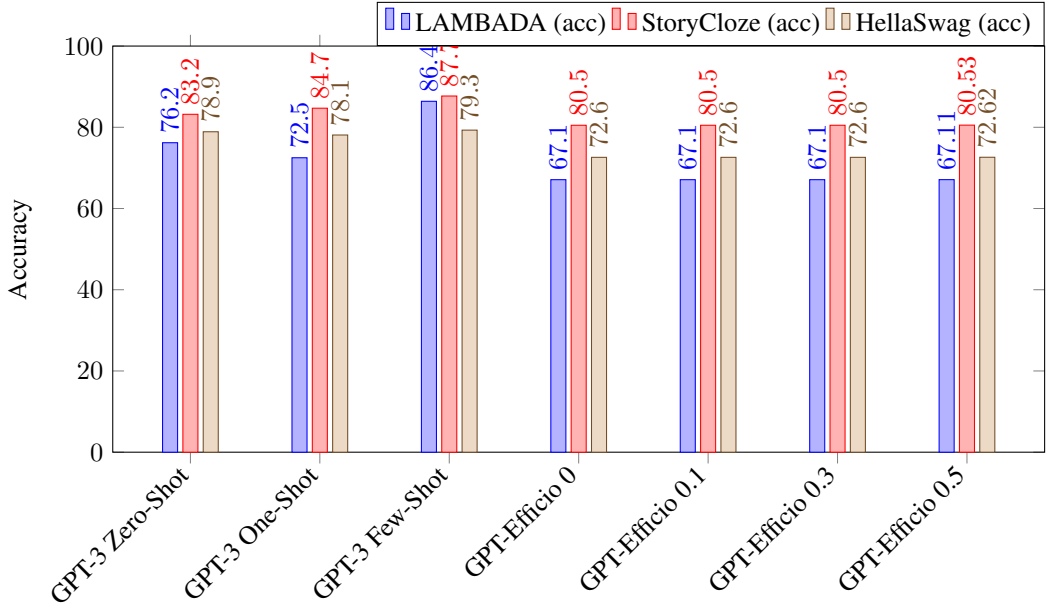
\begin{figure}
\begin{tikzpicture}
\begin{axis}[
    ybar, ymin=0, ymax=100,
    bar width=.2cm,
    width=\textwidth,
    height=.5\textwidth,
    legend style={at={(0.5,-0.15)},
      anchor=north,legend columns=-1},
    ylabel={Accuracy},
    symbolic x coords={GPT-3 Zero-Shot, GPT-3 One-Shot, GPT-3 Few-Shot, GPT-Efficio 0, GPT-Efficio 0.1, GPT-Efficio 0.3, GPT-Efficio 0.5},
    xtick=data,
    x tick label style={rotate=45,anchor=east},
    ymin=0,
    enlarge x limits=0.1,
    nodes near coords,
    nodes near coords align={vertical},
    every node near coord/.append style={rotate=90, anchor=west},
    legend style={at={(1,1)},
        anchor=south east,legend columns=-1}
    ]
\addplot coordinates {
    (GPT-3 Zero-Shot, 76.2)
    (GPT-3 One-Shot, 72.5)
    (GPT-3 Few-Shot, 86.4)
    (GPT-Efficio 0, 67.1)
    (GPT-Efficio 0.1, 67.1)
    (GPT-Efficio 0.3, 67.1)
    (GPT-Efficio 0.5, 67.11)
};
\addlegendentry{LAMBADA (acc)}
\addplot coordinates {
    (GPT-3 Zero-Shot, 83.2)
    (GPT-3 One-Shot, 84.7)
    (GPT-3 Few-Shot, 87.7)
    (GPT-Efficio 0, 80.5)
    (GPT-Efficio 0.1, 80.5)
    (GPT-Efficio 0.3, 80.5)
    (GPT-Efficio 0.5, 80.53)
};
\addlegendentry{StoryCloze (acc)}
\addplot coordinates {
    (GPT-3 Zero-Shot, 78.9)
    (GPT-3 One-Shot, 78.1)
    (GPT-3 Few-Shot, 79.3)
    (GPT-Efficio 0, 72.6)
    (GPT-Efficio 0.1, 72.6)
    (GPT-Efficio 0.3, 72.6)
    (GPT-Efficio 0.5, 72.62)
};
\addlegendentry{HellaSwag (acc)}
\end{axis}
\end{tikzpicture}
\caption{Analysis of the effects of hyperparameter synthetic to real data rate on completion tasks}
\end{figure}

\begin{table}[!htbp]
\centering
\small
\caption{Analysis of the effects of hyperparameter synthetic to real data rate on QA tasks}\label{syn-anal-qa}
\begin{tabular}{p{0.18\linewidth} p{0.06\linewidth} p{0.06\linewidth} p{0.15\linewidth} p{0.15\linewidth} p{0.12\linewidth}}
\toprule
\textbf{Model} & \textbf{syn\%} & \textbf{$n_{params}$} & \textbf{NQ} & \textbf{WebQ} & \textbf{TriviaQA}\\ 
\midrule
GPT-3 Zero-Shot & - & 175B & 14.6 & 14.4 & 64.3   \\ 
GPT-3 One-Shot & - & 175B & 23.0 & 25.3 & 68.0   \\ 
GPT-3 Few-Shot & - & 175B & 29.9 & 41.5 & 71.2   \\ 
GPT-Efficio & - & 950M & 27.5 & 40.6 & 69.2 \\
GPT-Efficio & .1 & 950M & 27.71 & 40.75 & 69.56 \\
GPT-Efficio & .3 & 950M & 28.68 & 41.70 & 70.35 \\
GPT-Efficio & .5 & 950M & 26.08 & 39.01 & 68.14 \\
\bottomrule
\end{tabular}
\end{table}

Table~\ref{syn-anal-qa} shows the GPT-Efficio performance with and without synthetic data in comparison with GPT-3 in question answering tasks. 

\begin{figure}
\begin{tikzpicture}
\begin{axis}[
    ybar, ymin=0, ymax=100,
    bar width=.2cm,
    width=\textwidth,
    height=.5\textwidth,
    legend style={at={(0.5,-0.15)},
      anchor=north,legend columns=-1},
    ylabel={Accuracy},
    symbolic x coords={GPT-3 Zero-Shot, GPT-3 One-Shot, GPT-3 Few-Shot, GPT-Efficio 0, GPT-Efficio 0.1, GPT-Efficio 0.3, GPT-Efficio 0.5},
    xtick=data,
    x tick label style={rotate=45,anchor=east},
    ymin=0,
    enlarge x limits=0.1,
    nodes near coords,
    nodes near coords align={vertical},
    every node near coord/.append style={rotate=90, anchor=west},
    legend style={at={(1,1)},
        anchor=south east,legend columns=-1}
    ]
\addplot coordinates {
    (GPT-3 Zero-Shot, 14.6)
    (GPT-3 One-Shot, 23.0)
    (GPT-3 Few-Shot, 29.9)
    (GPT-Efficio 0, 27.5)
    (GPT-Efficio 0.1, 27.71)
    (GPT-Efficio 0.3, 28.68)
    (GPT-Efficio 0.5, 26.08)
};
\addlegendentry{NQ}
\addplot coordinates {
    (GPT-3 Zero-Shot, 14.4)
    (GPT-3 One-Shot, 25.3)
    (GPT-3 Few-Shot, 41.5)
    (GPT-Efficio 0, 40.6)
    (GPT-Efficio 0.1, 40.75)
    (GPT-Efficio 0.3, 41.70)
    (GPT-Efficio 0.5, 39.01)
};
\addlegendentry{WebQ}
\addplot coordinates {
    (GPT-3 Zero-Shot, 64.3)
    (GPT-3 One-Shot, 68.0)
    (GPT-3 Few-Shot, 71.2)
    (GPT-Efficio 0, 69.2)
    (GPT-Efficio 0.1, 69.56)
    (GPT-Efficio 0.3, 70.35)
    (GPT-Efficio 0.5, 68.14)
};
\addlegendentry{TriviaQA}
\end{axis}
\end{tikzpicture}
\caption{Analysis of the effects of hyperparameter synthetic to real data rate on QA tasks}
\end{figure}
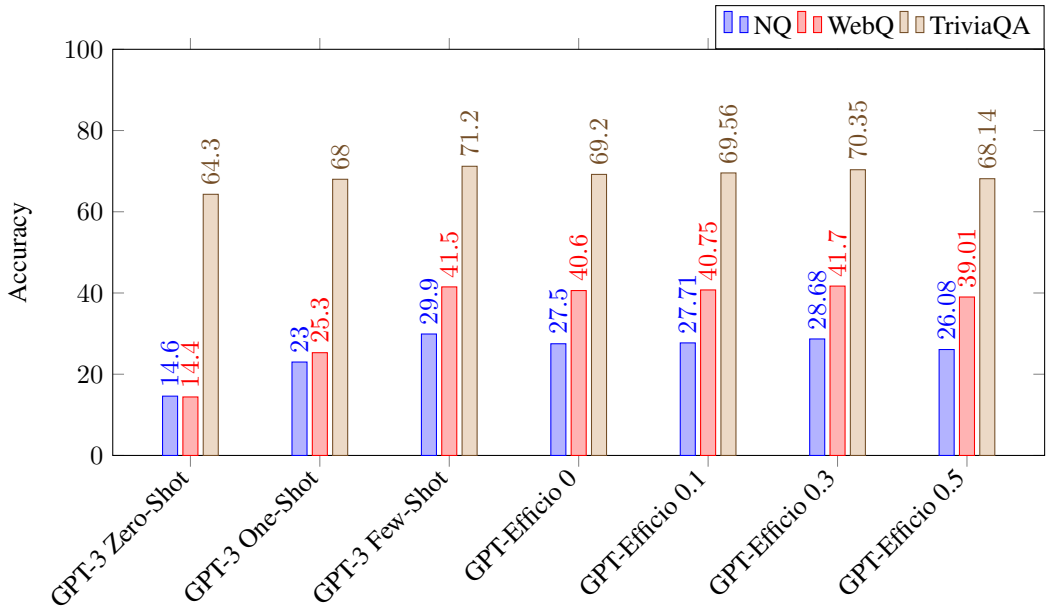

\section{Limitations}
The approach of synthetic data generation in Natural Language Processing (NLP), particularly using template-based question generation, does come with certain limitations that can impact its effectiveness. Here are some key limitations to consider:

\begin{enumerate}
    \item Quality of Synthetic Data: One of the biggest challenges is ensuring that the synthetic data generated is of high quality and closely mirrors the statistical properties of real-world data. If the synthetic data is of poor quality or does not accurately reflect the kinds of questions and answers the model will encounter in real-world situations, it can negatively impact the model's performance.

    \item Limited Diversity: Template-based question generation relies on predefined question templates. While this approach can produce a wide range of questions, it's inherently limited by the number and types of templates used. This method may not capture all possible ways of phrasing questions or handling complex sentence structures, which can limit the diversity of the generated questions.

    \item Lack of Nuance: Template-based generation can struggle to capture the nuances of natural language, particularly for complex sentences or subtleties in meaning. This is because it uses a relatively rigid, rule-based method to create questions, which can fail to account for context-dependent nuances in how questions might be phrased.

    \item Risk of Overfitting: There's a risk that the model will overfit to the patterns in the synthetic data, especially if a high ratio of synthetic to real data is used. This can lead to the model performing poorly on real-world data, as it may have learned patterns that are not representative of real-world language use.

    \item Computational Costs: Generating synthetic data, especially on a large scale, can be computationally intensive and time-consuming. This might not be an issue for smaller tasks or when using powerful hardware, but for larger tasks or resource-constrained situations, it could be a significant limitation.

    \item Annotation Quality: If synthetic data generation includes an annotation process (for instance, automatically generating labels for synthetic data), the quality of these annotations is crucial. Errors in annotation can introduce noise into the training data, which can negatively impact the model's performance.
\end{enumerate}

While these limitations pose challenges, they can be mitigated by using synthetic data generation in conjunction with other techniques. For instance, combining template-based question generation with more flexible, machine-learning-based methods can help to generate a wider variety of questions. Also, fine-tuning the model on real-world data after initial training on synthetic data can help to avoid overfitting. Ultimately, the careful design of the synthetic data generation process and rigorous validation of model performance are key to effectively using this approach.

\section{Future Work}
The approach of synthetic data generation in Natural Language Processing (NLP) has shown promise, but there's still much room for improvement and exploration. Here are some potential directions for future work:

\begin{itemize}
    \item Improving Synthetic Data Quality: One of the main challenges with synthetic data is ensuring its quality. Future work could focus on developing new techniques to generate higher-quality synthetic data that more accurately reflects real-world language patterns and distributions. 

    \item Hybrid Generation Methods: Combining template-based question generation with more flexible methods, such as machine learning or transformer-based question generation techniques, could create a more diverse set of synthetic questions and mitigate some of the limitations of template-based generation.

    \item Evaluation Metrics for Synthetic Data: Designing metrics to evaluate the quality of synthetic data could be a valuable contribution. These metrics could help guide the generation process and provide a more objective measure of whether the synthetic data is likely to improve model performance.

    \item Adaptive Synthetic Data Generation: Research could be directed towards adaptive synthetic data generation, where the synthetic data generation process is guided by the performance of the model, focusing on areas where the model struggles.

    \item Investigating Optimal Ratios of Synthetic to Real Data: More extensive empirical studies could help identify the optimal ratios of synthetic to real data for various types of NLP tasks and models. 

    \item Application-Specific Synthetic Data: Different NLP tasks might benefit from different types of synthetic data. Future work could investigate how to tailor synthetic data generation to specific applications.

    \item Addressing Biases: Future work could also focus on how synthetic data generation can be used to mitigate biases in NLP models, exploring different strategies for generating synthetic data that helps to counteract known biases in the training data.

    \item Computational Efficiency: Reducing the computational cost of synthetic data generation is another important direction for future work. This could involve developing more efficient algorithms or making better use of hardware resources.
\end{itemize}

By pursuing these avenues of future work, the field can continue to advance the use of synthetic data in NLP and fully realize its potential for improving the performance and robustness of language models.

\section{Conclusion}
The realm of Natural Language Processing (NLP) stands at an intriguing crossroads, with synthetic data generation emerging as a powerful ally in addressing data scarcity and model generalization challenges. Our exploration of template-based question generation has elucidated both its potential and the caveats that accompany its use. The augmentation cbilities it brings to the table can significantly bolster model training, especially in scenarios where real-world annotated datasets are sparse. Yet, the inherent rigidity of templates and the potential for overfitting demand a judicious and well-calibrated approach.

The interplay between synthetic and real-world data is a delicate balance. As demonstrated, the ratio between the two can substantially influence a transformer model's performance, emphasizing the necessity for meticulous empirical tuning. Moreover, while template-based strategies offer streamlined data generation, they ought to be integrated with other synthetic data techniques to ensure comprehensive model training.

Looking forward, as NLP continues its trajectory of rapid innovation, synthetic data generation's role will undoubtedly evolve. Researchers and practitioners should remain cognizant of the ever-shifting dynamics between real and synthetic data. Continuous evaluation, adaptive strategies, and openness to hybrid methodologies will be the bedrock upon which the next wave of NLP breakthroughs will be founded. The journey of integrating synthetic data in NLP is replete with both challenges and opportunities, beckoning the community to navigate its complexities with discernment and creativity.

\bibliographystyle{plainnat}
\bibliography{main}

\end{document}